\documentclass[sigconf,screen]{acmart}

\AtBeginDocument{%
  }

\acmConference[ACM-BCB '24]{15th ACM International Conference on Bioinformatics, Computational Biology and Health Informatics}{November 22, 2024}{Shenzhen, Guangdong Province, P.R. China}

\RequirePackage{graphicx}
 \usepackage{booktabs}
\usepackage{enumitem}
\usepackage{graphicx}

\usepackage{bm}

\usepackage{longtable}
\makeatletter
\def\set@curr@file#1{\def\@curr@file{#1}} 
\makeatother
\usepackage[load-configurations=version-1]{siunitx} 

\usepackage[mathlines, switch]{lineno}

\usepackage{booktabs}

\usepackage{graphicx} 
\usepackage{booktabs} 
\usepackage{multirow} 

\newcommand{\fullname}{Clinical Trial Multi-Agent System}
\newcommand{\mname}{ClinicalAgent}

\title[\mname: Clinical Trial Multi-Agent System]{\mname: \fullname~with Large Language Model-based Reasoning}

\author{Ling Yue}
\email{yuel2@rpi.edu}
\affiliation{%
  \institution{Rensselaer Polytechnic Institute}
  \city{Troy}
  \state{NY}
  \country{USA}
}

\author{Sixue Xing}
\email{xings@rpi.edu}
\affiliation{%
  \institution{Rensselaer Polytechnic Institute}
  \city{Troy}
  \state{NY}
  \country{USA}
}

\author{Jintai Chen}
\email{cjt147@illinois.edu}
\affiliation{%
  \institution{University of Illinois Urbana-Champaign}
  \city{Chempaign}
   \state{IL}
  \country{USA}
}

\author{Tianfan Fu}
\email{fut2@rpi.edu}
\affiliation{%
 \institution{Rensselaer Polytechnic Institute}
 \city{Troy}
 \state{NY}
 \country{USA}
 }

\begin{document}

\begin{abstract}
Large Language Models (LLMs) and multi-agent systems have shown impressive capabilities in natural language tasks but face challenges in clinical trial applications, primarily due to limited access to external knowledge. 
Recognizing the potential of advanced clinical trial tools that aggregate and predict based on the latest medical data, we propose an integrated solution to enhance their accessibility and utility. 
We introduce Clinical Agent System (\mname), a clinical multi-agent system designed for clinical trial tasks, leveraging GPT-4, multi-agent architectures, LEAST-TO-MOST, and ReAct reasoning technology. 
This integration not only boosts LLM performance in clinical contexts but also introduces novel functionalities. 
The proposed method achieves competitive predictive performance in clinical trial outcome prediction (0.7908 PR-AUC), obtaining a 0.3326 improvement over the standard prompt Method. 
Publicly available code can be found at \url{https://anonymous.4open.science/r/ClinicalAgent-6671}.
\end{abstract}

\ccsdesc[100]{Computing methodologies~Discourse, dialogue and pragmatics}
\ccsdesc[500]{Computing methodologies~Multi-agent planning}
\ccsdesc[100]{Computing methodologies~Supervised learning by classification}

\keywords{Clinical Trial, Clinical Trial Outcome Prediction, Drug Development, Multi-Agent Planning, Large Language Models, Large Language Model-based Reasoning, Healthcare}

\maketitle

\section{Introduction}




The introduction of clinical multi-agent systems into the healthcare sector marks a substantial advancement in improving care quality through sophisticated computational methods and in-depth data analysis. 
Modern medicine increasingly relies on advanced technologies to enhance patient outcomes, streamline clinical processes, and provide deeper insights into complex health conditions. These advancements are driven by the integration of Large Language Models~\citep{singhal2023large} (LLMs) like ChatGPT~\citep{liu2023utility}, BioGPT~\citep{luo2022biogpt}, ChatDoctor~\citep{yunxiang2023chatdoctor}, 
and Med-PaLM~\citep{singhal2023towards}, which have shown considerable success in processing and understanding medical data, providing customized care, and offering insights into intricate health conditions. 
However, their use in clinical trials faces challenges, mainly due to their limited ability to access and integrate external knowledge sources, such as DrugBank~\citep{wishart2018drugbank}. 
This research stems from the urgent need to fully utilize LLMs in clinical settings, going beyond the conversational skills of current models to include actionable and explanatory analysis leveraging extensive external data.

Our study introduces \mname, a new Clinical multi-agent system tailored for clinical trial tasks. Utilizing the capabilities of GPT-4, combined with multi-agent system architectures, and incorporating advanced reasoning technologies like LEAST-TO-MOST~\citep{zhou2023leasttomost} and ReAct~\citep{yao2023react}, our solution not only boosts LLM performance in clinical scenarios but also brings new functionalities. Our system is designed to autonomously oversee the clinical trial process, filling the void in existing implementations that mainly focus on conversational interactions without sufficient actionable outcomes.

Prior research has highlighted the potential of LLMs in healthcare, particularly in diagnostics, patient communication, and medical research~\citep{singhal2023towards,yunxiang2023chatdoctor,singhal2023large}. Yet, these investigations have not fully exploited the models for clinical trials, where understanding the complex relationships between drugs, diseases, and patient reactions is crucial~\citep{lu2019integrated,lu2023genocraft}. Our research introduces a multi-agent framework that uses specialized agents for tasks such as drug information retrieval, disease analysis, and explanatory reasoning. This strategy not only allows for a more detailed and understandable decision-making process but also significantly enhances clinical trial analysis capabilities, including predicting outcomes, deciphering reasons for failure~\citep{chen2024uncertainty}, and estimating trial duration~\citep{yue2024trialdura}.

A review of the literature indicates a growing interest in improving LLM applications in medicine. 
For instance, studies like \citep{li2024chatgpt} discuss employing ChatGPT and BioGPT for patient data synthesis and diagnostic recommendations. 
However, these discussions often focus only on the conversational aspects, overlooking the actionable intelligence and comprehensive reasoning our approach introduces. 
Moreover, our method is unique in incorporating external databases and reasoning technologies like ReAct, aiming not just to interpret but also to act on the intricate network of clinical data.

Our main contributions are summarized as follows:
\begin{itemize}
\item We present \fullname~(\mname), the first multi-agent framework that elevates the conversational abilities of LLMs with actionable intelligence.
\item We integrate extensive tools, and knowledge and use advanced reasoning technologies to enhance the system's decision-making capabilities.
\item \mname~achieves competitive predictive performance in clinical trial outcome prediction (0.7908 PR-AUC), obtaining a 0.3326 improvement over the standard large language model using prompt. 
\end{itemize}

In the following sections, we discuss the related works, explore the methodology, detail the structure and functionality of our multi-agent system, outline our experimental strategy, and discuss the significance of our findings within the clinical trial context.

\section{Related Work}
Natural Language Processing (NLP) has achieved significant progress in the biomedical arena, delivering crucial insights and tools for a range of applications in healthcare and medicine. The advent of Large Language Models (LLMs) has notably advanced the medical field by embedding comprehensive medical knowledge into their training. 

For example, question answering (QA) in the medical domain represents a critical challenge in NLP, where language models are tasked with responding to specific queries using their embedded medical knowledge, \textit{e.g.,} MedQA (USMLE)~\citep{jin2021disease} 
HeadQA~\citep{headqa}, MMLU~\citep{mmlu}, and PubMedQA~\citep{pubmedqa}.

Despite being pre-trained for general purposes, closed-source LLMs like ChatGPT \citep{openai2022chatgpt} and GPT-4 \citep{openai2023gpt4} have demonstrated considerable medical capabilities in both benchmark evaluations and real-world applications. \cite{liévin2023large} applied GPT-3.5 using various prompting techniques, such as Chain-of-Thought, few-shot, and retrieval augmentation, across three medical reasoning benchmarks, showcasing the model’s robust medical reasoning skills without the need for specialized fine-tuning. 
Additionally, evaluations of LLMs such as ChatGPT on professional medical assessments, including the US Medical Exam \citep{kung2023performance} and the Otolaryngology-Head and Neck Surgery Certification Examinations \citep{Long2023ANE}, have resulted in scores that meet or nearly meet passing thresholds. This performance underscores the potential of LLMs to aid in significant medical contexts, including medical education and clinical decision-making.

\paragraph{AI for Clinical Trial.} 
AI has great potential to revolutionize clinical trials and, more generally, the biomedical industry in a couple of problems. 
Specifically, 
TrialBench~\cite{chen2024trialbench} highlights eight critical issues in clinical trials that are ready for AI solutions, including the forecast of trial duration, estimation of patient dropout rates, prediction of serious adverse events, mortality rates, outcomes of trial approvals, identification of trial failure reason, optimization of drug dosing, and the design of eligibility criteria. 
\cite{zhang2020deepenroll,gao2020compose} leverage AI to recruit appropriate patients that meet the requirement in eligibility criteria. 
\cite{fu2022hint,fu2023automated,lu2024uncertainty} builds machine learning models to predict the outcome of clinical trials based on clinical trial features such as drug molecule, disease code, and eligibility criteria. 
To predict clinical trial enrollment success, \cite{yue2024trialenroll} leverages large language model-augmented features and customized deep \& cross network to model the text feature. 
Multi-omics data enables finer-grained analysis, which is crucial for precision medicine approaches (personalized therapy). By understanding the genetic~\cite{lu2021cot,lu2022cot,zhang2021ddn2}, transcriptomic~\cite{lu2023deep,fu2024ddn3}, and other molecular profiles of patients~\cite{chen2021data}, treatments can be customized to match individual disease mechanisms/biological pathway~\citep{wu2022cosbin}, potentially leading to more effective and personalized therapies~\citep{yi2018enhance}. 
In the context of clinical trials, \citet{wang2024twin} leverages the large language model to generate patient-level digital twins to simulate clinical trials. 
However, most of these works do not utilize LLM's reasoning ability and cannot solve complex problems in clinical trials. 
To the best of our knowledge, our work is the first LLM agent work for clinical trials and is able to solve complex clinical trial reasoning problems.

\section{Methods}

\begin{figure*}
\centering
\includegraphics[width=0.8\textwidth,keepaspectratio]{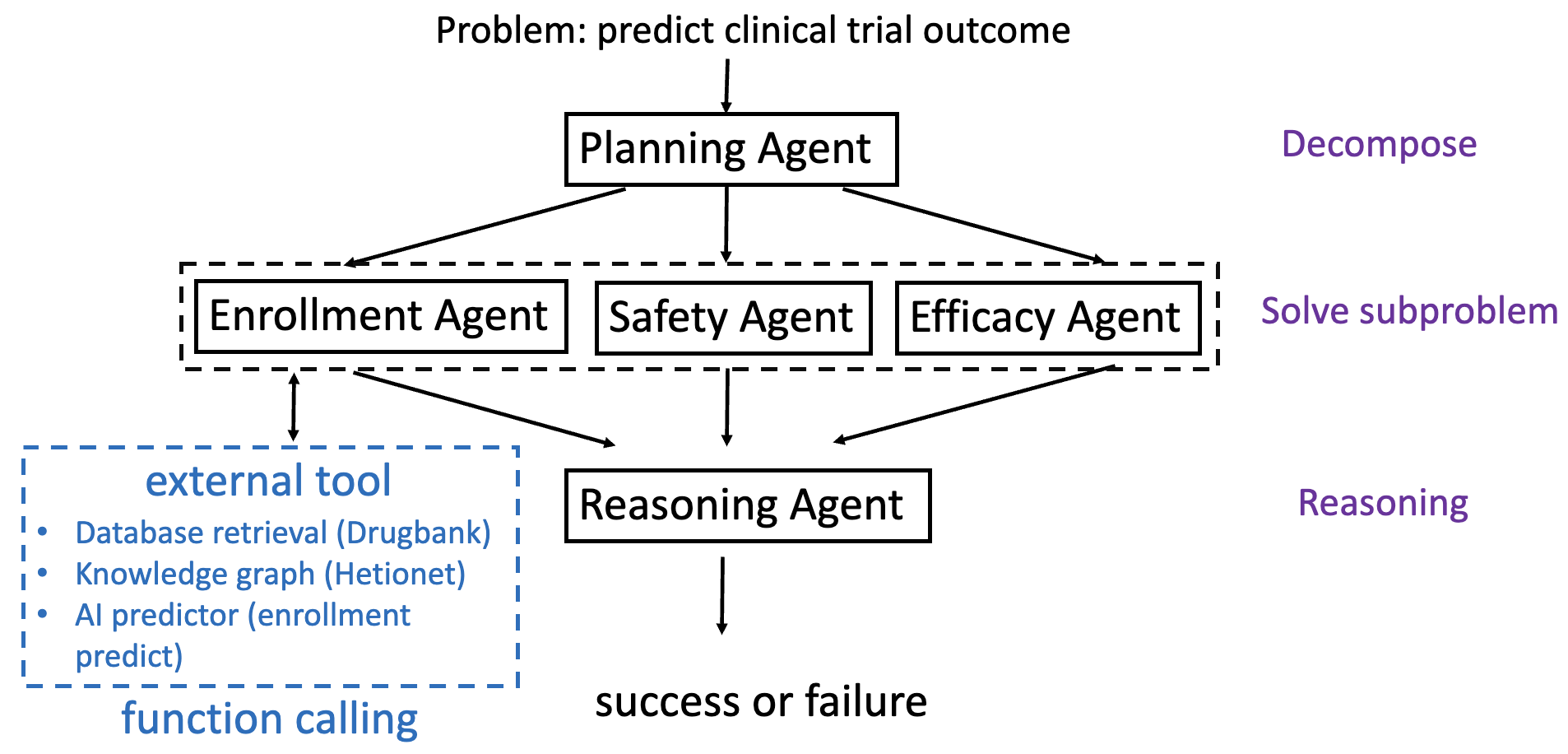}
\caption{\mname~framework. Given a complex problem to solve (\textit{e.g.,} predicting clinical trial outcome), the role of the Planning Agent is to decompose it into three subproblems: trial enrollment, drug safety to the human body, and drug efficacy to disease. These subproblems are solved by Enrollment Agent, Safety Agent, and Efficacy Agent, respectively, enhanced by calling external tools (Section~\ref{sec:external_tool}). Finally, the Reasoning Agent aggregates the solutions of subproblems, draws the conclusion, and makes the prediction. }
\label{fig:overview}
\end{figure*}

\subsection{Overview of \mname}
Our proposed system is a conversational multi-agent framework analogous to a hospital staffed by various specialists. 
Each agent within this system plays a distinct role, 
mirroring the specialization seen in medical professionals—some focus on pharmacology, 
others on diagnosing diseases, while a few are dedicated to designing clinical trials. 
To process natural language inputs and generate responses that are coherent and contextually appropriate, 
each agent utilizes GPT-4. 
Moreover, we enhance the system's reasoning capabilities by incorporating methodologies such as ReAct~\citep{yao2023react} and the LEAST-TO-MOST~\citep{zhou2023leasttomost} principle. 
Following the reasoning process, the system is capable of taking actions such as searching for information, 
indexing data in databases, and employing expert AI models. 
By integrating this information, the system effectively simulates a highly knowledgeable doctor. 
Working in concert, these agents can deliver precise, explainable solutions to user inquiries.

\mname~integrates diverse machine learning models and data sources to predict clinical trial duration, failure reasons, outcomes, and enrollment difficulty.
As the first multi-agent framework for clinical trials using advanced LLM technology, it features a developed website and aims to become a community platform, providing precise and explainable solutions to user inquiries.
In the following paragraph, we will introduce an example of how to use \mname~to predict clinical trial outcomes.

\subsection{Agent Roles and Responsibilities}
The \mname~framework integrates a diverse array of specialized agents, each employing the ReAct and LEAST-TO-MOST reasoning methods to meticulously plan their actions. 
Through the use of advanced search capabilities, access to specialist models, and indexing in databases, these agents are able to execute a wide range of tasks effectively. 
Below, we delve into the specific roles and responsibilities assigned to each agent within the system.

\subsubsection{Planning Agent}
The Planning Agent's primary role is to strategize and determine the optimal approach to address user problems. Utilizing the LEAST-TO-MOST Reasoning method, this agent systematically decomposes complex issues into smaller, more manageable subproblems. This stepwise breakdown facilitates targeted interventions, where each subproblem is addressed by the most suitable specialist agent. In the context of clinical trials, the Planning Agent employs few-shot learning techniques to train on example scenarios. This approach enhances the agent's ability to effectively decompose and delegate tasks within clinical contexts, ensuring precise and efficient problem-solving.

\subsubsection{Efficacy Agent}
The Efficacy Agent is a specialized module within our multi-agent framework, primarily focused on assessing the therapeutic effectiveness of drugs against specified diseases. This agent utilizes advanced data retrieval and analysis techniques, drawing from rich biomedical databases such as DrugBank~\citep{wishart2018drugbank} and the HetioNet Knowledge Graph to ensure comprehensive and accurate evaluations.

Specifically, the Efficacy Agent employs the SMILES (Simplified Molecular Input Line Entry System) notation to identify and retrieve detailed chemical and pharmacological information about drugs. This includes their molecular structure, mechanism of action, metabolism, and potential side effects, providing a holistic view of the drug's properties. 

Upon receiving a query with a specific drug and disease, the Efficacy Agent performs several key functions:
\begin{itemize}
    \item \textbf{Drug and Disease Profiling:} Retrieves up-to-date, detailed descriptions of the drug and the disease from DrugBank and other relevant databases, ensuring that users have access to reliable and comprehensive information.
    \item \textbf{Interaction Pathway Mapping:} Utilizes the HetioNet Knowledge Graph to trace and visualize the pathways connecting the drug to the disease. This involves identifying biological interactions, such as target proteins and genetic associations, that are crucial for understanding the drug's potential efficacy.
    \item \textbf{Efficacy Assessment:} Analyzes the gathered information to evaluate the potential effectiveness of the drug against the disease, considering factors like target specificity, therapeutic indices, and evidence from clinical trials.
\end{itemize}

By synthesizing data from multiple sources and employing sophisticated analytical techniques, the Efficacy Agent provides essential insights into drug-disease relationships, supporting informed decision-making in clinical and research settings.

\subsubsection{Safety Agent}
The Safety Agent is integral to our \mname~framework, focusing specifically on the assessment of drug safety and its implications for patient health. This agent leverages a comprehensive repository of pharmacological data and historical clinical trial outcomes to evaluate the risks associated with specific drug-disease interactions. Utilizing databases such as DrugBank and clinical trial registries, the Safety Agent provides detailed insights into the historical safety profiles of drugs.

Key functions of the Safety Agent include:
\begin{itemize}
    \item \textbf{Drug Safety Profiling:} Accesses detailed safety information from databases to compile historical data on adverse drug reactions, contraindications, and warnings. This data is crucial for understanding the risk factors associated with the drug.
    \item \textbf{Historical Failure Rate Analysis:} Investigates past clinical trials and reported outcomes to determine the failure rates of drugs in similar contexts or against similar diseases. This analysis helps predict potential safety concerns in current applications.
    \item \textbf{Risk Assessment:} Employs statistical models to analyze the safety data and predict the risk of adverse effects when a drug is used to treat a particular disease. This predictive capability is vital for making informed decisions about drug prescriptions and usage.
\end{itemize}

By systematically analyzing safety data and historical trial outcomes, the Safety Agent plays a crucial role in minimizing risks and enhancing patient safety in clinical settings.

\subsubsection{Enrollment Agent}

Proper enrollment ensures that the trial has enough participants to statistically power the study. This is essential to detect the true effect of the intervention being tested. Insufficient enrollment can lead to inconclusive or unreliable results because the sample size determines the ability of a trial to accurately reflect the effects of a treatment. 
We design a Hierarchical transformer-based model that takes eligibility criteria as an input feature and predicts the success rate of enrollment. It is a binary classification problem, where 1 denotes the successful enrollment while 0 does not. The details of the model can be found in Section~\ref{app:enroll} in the Appendix.

\subsection{Calling External Tools}
\label{sec:external_tool}

GPT supports calling external tools (\textit{e.g.,} function, database retrieval) to leverage external knowledge and enhance its capability. 
Specifically, suppose we have a couple of toolkits. GPT's API can automatically detect which tool to use, which serves as glue to connect large language models to external tools.
Our system integrates a variety of external data sources and predictive AI models to support the agents' functions. 

\paragraph{Data Sources}
The use of professional datasets is pivotal in ensuring the accuracy and reliability of our agents' information retrieval capabilities.
\begin{itemize}
    \item \textbf{Drug Databases:}
      Drugbank~\citep{wishart2018drugbank} stands out as a premier resource, offering detailed drug data, including chemical, pharmacological, and pharmaceutical information, with a focus on comprehensive drug-target interactions.
      Drugbank is not only a repository of drug information but also serves as an invaluable tool for bioinformatics and cheminformatics research, providing data for over 13,000 drug entries including FDA-approved small molecule drugs, FDA-approved biopharmaceuticals (proteins, peptides, vaccines, and allergenics), and nutraceuticals. \\
    \item \textbf{Knowledge Graphs:}
      Hetionet~\citep{himmelstein2017systematic} is an integrative network of biology that encompasses a comprehensive collection of biological entities and their relationships.
      It uniquely combines data from various biomedical databases covering diseases, genes, compounds, and more, into a single, coherent graph structure.
      This interconnected approach allows for multifaceted analyses, including drug repurposing, genetic associations, and network medicine.
      Hetionet includes over 47,000 nodes of different types (\textit{e.g.,} diseases, drugs, genes) and more than 2 million relationships, offering a rich dataset for computational biology and drug discovery. \\
    \item \textbf{ClinicalTrials.gov:} We extracted data from \url{https://clinicaltrials.gov/}, which includes information from both completed and ongoing clinical trials. This data is essential for validating our predictive models and for training them to understand clinical outcomes based on past trial data. \\
    \item \textbf{LLM-generated data:}
      Large Language Models (LLM) like GPT-4 and its successors, have demonstrated remarkable capability as knowledge compressors and generators.
      They can synthesize and extrapolate information from vast datasets to generate coherent, novel data points and insights.
      In this research, we leverage LLMs to generate new knowledge relevant to our study, including hypothetical drug interactions, potential therapeutic targets, and model organism analyses.
      This approach allows us to expand our dataset beyond traditional sources, incorporating generated insights that are validated against existing databases and literature.
      The use of LLM-generated data introduces a novel dimension to our research, enabling the exploration of uncharted territories in drug discovery and biomedical research. 
\end{itemize}

\paragraph{Predictive AI Models} 

We utilize multiple predictive AI models within our framework to ensure the accuracy and reliability of our agents' abilities:

\begin{itemize}
    \item \textbf{Enrollment Model:}
The enrollment model is designed to predict the likelihood of successful participant enrollment in clinical trials based on the eligibility criteria, the drugs involved, and the diseases targeted~\citep{yue2024trialenroll}. This is a hierarchical transformer-based model, integrating sentence embeddings from BioBERT~\citep{lee2020biobert} to capture the nuanced medical semantics in the criteria text. In practice, the Enrollment Agent receives a query containing the drugs, diseases, and detailed eligibility criteria. It processes this information to predict the enrollment difficulty, which aids in planning and adjusting recruitment strategies for clinical trials. This capability supports more efficient trial design and can significantly impact the speed and success of new drug developments. The details can be found in Section~\ref{app:enroll} in Appendix.

    \item \textbf{Drug Risk Model:}
        The Drug Risk Model is designed to estimate the likelihood of a drug not achieving the desired therapeutic effect in clinical trials. This model is based on historical data of drug performances across various trials. Using a simple but effective approach, each drug is represented by its historical success rate, calculated as the mean of its trial outcomes (1 for success and 0 for failure).
        
        We store these success rates in a precomputed dictionary and utilize a lookup mechanism to assess drug risk rapidly. For drugs not found in the dictionary, a matching function approximates the closest drug name to ensure robust risk assessments. This method allows for quick and accurate risk estimations in real-time decision-making processes and is particularly useful in early-stage drug development and trial planning.

    \item \textbf{Disease Risk Model:}
        Parallel to the Drug Risk Model, the Disease Risk Model calculates the probability of unsatisfactory treatment outcomes associated with specific diseases. This model aggregates historical trial data to determine success rates for diseases, which are then inverted to represent risk levels.
        
        Similar to the drugs model, each disease's risk is precomputed and stored. The model employs sophisticated string-matching techniques to accommodate variations in disease naming conventions, ensuring accurate risk evaluations. This model aids in the prioritization of diseases in clinical research and helps in forecasting the challenges in achieving successful treatment outcomes.    
\end{itemize}

\subsection{Integration of Reasoning Technology}
To further enhance the agent's decision-making capabilities, we integrate advanced reasoning technologies such as ReAct (recognition, action, and context)~\citep{yao2023react} and the Least-to-Most reasoning framework~\citep{zhou2023leasttomost}. These methodologies complement each other by providing robust mechanisms for addressing complex problems through structured and contextual analysis.

\textbf{ReAct Reasoning:}
ReAct reasoning is a holistic approach that emphasizes the critical roles of recognition (Re), action (A), and context (Ct) in effective problem-solving. This methodology advocates for the identification of patterns or cues (recognition), the formulation and execution of a course of action (action), and the careful consideration of the surrounding circumstances (context). By integrating these elements, ReAct equips agents to make informed and precise decisions rapidly, an asset, particularly in dynamic and unpredictable environments.

\textbf{Least-to-Most Reasoning:}
In contrast, the Least-to-Most reasoning method adopts a hierarchical approach to problem-solving. It suggests beginning with the simplest or least complex aspects and gradually progressing to address more intricate components. This structured problem-solving sequence ensures that foundational elements are thoroughly understood before advancing to tackle more complex layers of the issue. This method is valuable in educational contexts and when dealing with new or unfamiliar concepts, promoting a comprehensive understanding and preventing potential oversights.

\textbf{Synergistic Integration:}
By combining ReAct and Least-to-Most reasoning, we can formulate a synergistic strategy that leverages the strengths of both methods. Initially, the Least-to-Most framework decomposes a problem into its elemental parts, organizing them from simplest to most complex. Subsequently, within this structured framework, ReAct reasoning is applied to each segment. This involves recognizing relevant patterns or cues, deciding on appropriate actions based on these insights, and adapting these actions by considering the immediate context. This integrative approach not only ensures a methodical breakdown of problems but also adopts solutions dynamically to meet the specific demands of each scenario.

\subsection{Workflow}

The workflow of our \mname~system is designed to optimize the collaboration and efficiency of multiple specialized agents to address complex medical inquiries. The process is structured in several sequential steps, as described below:

\paragraph{Step 1: Initial Planning and Problem Decomposition}
The workflow begins with the Planning Agent, which takes the lead in assessing the user's query. Utilizing the LEAST-TO-MOST Reasoning method, this agent decomposes the complex problem into simpler, more manageable subproblems. This structured breakdown is crucial as it allows for targeted problem-solving by directing specific tasks to the most appropriate specialist agents.

\paragraph{Step 2: Task Allocation to Specialist Agents}
Once the problem is decomposed, the Planning Agent allocates each subproblem to the respective specialist agents. For example:
\begin{itemize}
    \item The Efficacy Agent is tasked with assessing drug effectiveness against specific diseases.
    \item The Safety Agent evaluates potential risks and adverse effects associated with the drug.
    \item The Enrollment Agent handles the feasibility and strategies for patient enrollment in clinical trials.
\end{itemize}
Each agent operates independently, utilizing its specialized models and databases to process and analyze the assigned task.

\paragraph{Step 3: Independent Agent Processing}
Each specialist agent processes its assigned subproblems using specific methodologies and external tools. This includes retrieving and analyzing data from sources like DrugBank and HetioNet, applying predictive models, and generating insights based on the agent’s specialty. The agents may also call external functions or databases to enhance their assessments or predictions.

\paragraph{Step 4: Synthesis of Findings}
After each agent completes its task, the results are sent back to the Planning Agent. This agent synthesizes the findings from all the specialists, creating a comprehensive response that integrates all aspects of the problem, from drug efficacy and safety to enrollment potential.

\paragraph{Step 5: Reasoning and Final Decision Making}
The final step involves applying the ReAct reasoning method to the synthesized findings. Here, the Planning Agent, enhanced by few-shot learning capabilities, examines the context and details of the integrated response to make informed decisions. This approach ensures that the final recommendation or solution is not only based on segmented analysis but also considers the interdependencies and broader implications of the combined agent findings.

\paragraph{Step 6: Delivery of Solution}
The completed solution, which encompasses a detailed and reasoned response based on the collective intelligence of the multi-agent system, is then delivered to the user. This response not only addresses the initial query but also provides explanatory insights that justify the recommendations, thereby enhancing user trust and understanding.

This structured workflow ensures that \mname~effectively mimics a collaborative team of medical specialists, offering precise and comprehensive solutions to complex medical inquiries.

\section{Experiment}

This section outlines the experimental design used to assess the performance of \mname~in the setting of clinical trials. 
We aim to demonstrate that the \mname~framework significantly outperforms direct predictions by LLM. Additionally, compared to traditional machine learning models, \mname~also shows competitive results. It is important to note that all machine learning models can be integrated as part of \mname; this comparison between end-to-end ML models is merely to showcase the potential of \mname.

\subsection{Baseline Methods}

To ensure a comprehensive evaluation, we have selected diverse baseline methods known for their robustness in similar tasks:
\begin{enumerate}
    \item \textbf{Gradient-Boosted Decision Trees (GBDT)}: This method integrates embeddings for drugs, diseases, and eligibility criteria derived from BioBERT. The concatenated embeddings are then processed using LightGBM, a popular gradient-boosting framework that is highly efficient and scalable, making it suitable for handling complex datasets typical in clinical trials.
    \item \textbf{Hierarchical Attention Transformer (HAtten)}: Employing BioBERT embeddings for drugs and diseases, this model introduces a hierarchical attention mechanism. It systematically focuses on different granularity levels, from entire paragraphs to specific sentences within the eligibility criteria, enhancing its ability to discern relevant information. The process culminates in a two-layer Multilayer Perceptron (MLP), which aids in refining the decision process.
    \item \textbf{Standard Prompting}: As a control, this baseline employs large language models (LLMs) GPT-4~\citep{openai2022chatgpt} in their standard configuration. It tests the hypothesis that without tailored adaptations or integrations of external data, the pre-trained knowledge embedded within LLMs can competently perform outcome prediction in clinical trials, albeit potentially less effectively than more specialized approaches.
\end{enumerate}

This comparative analysis will help in highlighting the strengths and potential areas for improvement in \mname, guiding future enhancements in the model's architecture and its application in clinical trial settings.

\subsection{Experimental Setup}

Our experimental framework was implemented on a server equipped with an AMD Ryzen 9 3950X CPU, 64GB RAM, and an NVIDIA RTX 3080 Ti GPU. We utilized Python 3.8 for scripting and PyTorch for model implementation and training. For each experiment, we used the same seed to ensure reproducibility. For our experimental validation, we randomly selected 40 training samples from the clinical trial outcome prediction benchmark provided in \citep{fu2022hint,fu2023automated}. We applied the same approach to select 40 samples from the test set. The decision to use 40 samples was driven by the high computational cost associated with calling the OpenAI API, which necessitated a balance between thorough testing and resource constraints. By ensuring that our training and testing datasets were balanced and randomized, we aimed to provide a robust evaluation framework for our model's performance while managing the associated costs effectively.

\begin{table}[htbp]
\caption{A real example of \mname~on clinical trial outcome prediction. }
\label{tab:casestudy}
\begin{tabular}{p{\columnwidth}}
\toprule
\textbf{User Input} \\
\noindent\textbf{Problem}: I have designed a clinical trial and hope you can help me predict whether this trial can pass. 
Features contain (1) \textbf{drug}: Aggrenox capsule;
(2) \textbf{disease}: cerebrovascular accident;
(3) \textbf{inclusion criteria}: ...; (4) \textbf{exclusion criteria}: ...;
\\ 
\midrule
\noindent\textbf{Planning (Planning Agent)}: few-shot learning \\
\noindent\textbf{Thought}: Decompose a complex problem into several easier subproblems. \\ 
\noindent\textbf{Subproblem 1 (Enrollment Agent)}: Determine the level of enrollment feasibility based on inclusion and exclusion criteria. \\
\noindent\textbf{Subproblem 2 (Safety Agent)}: Evaluate the safety of the drug ``aggrenox capsule'' \\
\noindent\textbf{Subproblem 3 (Efficacy Agent)}: Assess the efficacy of the drug ``aggrenox capsule'' on the disease ``cerebrovascular accident'' \\ 
\midrule  
\textbf{Solving Subproblems}:  \\ 
\noindent\textbf{Subproblem 1 (Enrollment Agent)}: Detect the function ``enrollment prediction model'' should to be called. The predicted enrollment failure rate is 0.3597.  \\
\noindent\textbf{Subproblem 2 (Safety Agent)}: Detect functions ``disease historical statistics'' and ``drug historical statistics'' should be called. The historical failure rate of Aggrenox capsules in clinical trials is reported as 1.0. \\
\noindent\textbf{Subproblem 3 (Efficacy Agent)}: Detect functions ``retrieve Hetionet knowledge graph'' and ``retrieve DrugBank database'' should be called. \\ 
\midrule 
\textbf{Reasoning (Reasoning Agent)}: few-shot learning  \\
\noindent\textbf{1. Enrollment Difficulty (Enrollment Agent)}:  A moderate level of difficulty in enrolling participants. This could pose challenges in recruiting a sufficient number of eligible patients meeting the strict inclusion and exclusion criteria set for the trial, which can impact the overall success rate of the trial. \\ 
\noindent\textbf{2. Safety of Aggrenox Capsule (Safety Agent)}: The historical failure rate of Aggrenox capsules (1.0) indicates a lack of success in previous trials. Evaluating the safety profile of the Aggrenox capsule in the context of cerebral infarction and cerebrovascular accidents is crucial due to the specified risk factors and exclusion criteria. The presence of risk factors like hypertension, diabetes, and other conditions among the trial participants might raise concerns about potential interactions or adverse effects of Aggrenox, contributing to the historical failure rate. \\ 
\noindent\textbf{3. Efficacy of Aggrenox Capsule (Efficacy Agent)}: The Aggrenox capsule, combining aspirin and dipyridamole, is beneficial in preventing strokes by its antiplatelet and vasodilatory effects. Given the criteria provided for the clinical trial specifically targeting patients with a history of cerebral infarction who meet certain age and risk factor requirements, Aggrenox capsule's mechanism of action aligns with the needs of this patient population. \\ 
\textbf{Draw Conclusion}. Therefore, considering the historical failure rate, safety concerns, and the moderate enrollment difficulty level associated with the clinical trial design and the use of the Aggrenox capsule in patients with cerebral infarction, the predicted success rate of the trial is low, at 0.0. \textbf{(ground truth is 0)}
\\
\bottomrule
\end{tabular}
\end{table}

\begin{table*}[t!]
\small 
\centering
\caption{Predictive performance of various methods.}
\label{table:result}
\begin{tabular}{lcccccc}
\toprule 
Method & Accuracy ($\uparrow$) & ROC-AUC ($\uparrow$) & PR-AUC ($\uparrow$) & Precision ($\uparrow$) & Recall ($\uparrow$) & F1 ($\uparrow$) \\
\midrule
{GBDT} & 0.6250 & 0.8000 & 0.8669 & 0.6250 & \textbf{1.0000} & 0.7692 \\
{HAtten} & \textbf{0.7500} & 0.7573 & \textbf{0.8718} & \textbf{0.8947} & 0.6800 & \textbf{0.7727} \\
GPT-3.5 & 0.5250 & 0.4853 & 0.4419 & 0.4000 & 0.5333 & 0.4571 \\
GPT-4 & 0.6500 & 0.6800 & 0.4582 & 0.5385 & 0.4666 & 0.5000 \\ 
{\mname} & 0.7000 & \textbf{0.8347} & 0.7908  & 0.5714 & 0.8000 & 0.6667 \\
\bottomrule
\end{tabular}
\end{table*}

\begin{table*}[t!]
\centering
\caption{Impact of few-shot learning on \mname~performance.}
\label{table:few_shot_ablation}
\begin{tabular}{lcccccc}
\toprule
Method & Accuracy ($\uparrow$) & ROC-AUC ($\uparrow$) & PR-AUC ($\uparrow$) & Precision ($\uparrow$) & Recall ($\uparrow$) & F1 ($\uparrow$) \\ 
\midrule
\mname~w few-shot & 0.7 & 0.8347 & 0.7908 & 0.5714 & 0.8 & 0.6667 \\
\mname~w/o few-shot & 0.75 & 0.824 & 0.6793 & 0.647 & 0.7333 & 0.6875 \\
\bottomrule
\end{tabular}
\end{table*}

\subsection{Implementation Details}
In this section, we provide detailed descriptions of the implementation processes to enhance the reproducibility of our study.

\paragraph{Role Assignment to Agents}
Each agent within the \mname~framework is designated a specific role, which is integrated directly into the LLM’s system prompt for clarity and focus. For instance, the role of the Efficacy Agent is defined as follows: 
\begin{quote}
"As an efficacy expert, you have the capability to assess a drug's efficacy against diseases by examining its effectiveness on the disease."
\end{quote}
This role definition is crucial as it guides the LLM to prioritize responses based on the assigned expert domain, leveraging the model's inherent capability to focus more acutely on instructed tasks than on general information.

\paragraph{Defining External Tools}
External tools are defined in a structured format to facilitate their integration and usage within the LLM environment. These definitions are crafted in JSON format, specifying the function name, description, and necessary parameters. Examples can be found in the \ref{app:func}.

This structured approach allows for the direct transmission of function calls to the LLM, which in turn provides detailed responses including the function name and arguments. These responses enable the execution of functions locally and the retrieval of results in a structured manner.

\paragraph{Enhanced Few-Shot Reasoning}
To improve the model's reasoning capabilities, we incorporate examples of sub-problems and corresponding labels within the system prompt. This method, known as few-shot learning, aids the LLM in understanding the context and methodology required to solve complex problems by referencing similar, previously solved problems. This approach not only enhances the accuracy of the model’s outputs but also its ability to generalize from limited examples to new, unseen scenarios.

These implementation strategies collectively ensure that each component of \mname~operates effectively and that the integration between different agents and external tools is seamless, fostering an environment conducive to robust, reproducible research.

\subsection{Quantitative Results}

Table~\ref{table:result} presents the performance of various methods. Specifically, \mname~obtain the highest ROC-AUC score at 0.8347 among all the compared methods. We observe that \mname~achieve competitive performance among all the well-established methods, \textit{e.g.,} GBDT. 
Also, compared with the standard prompt (basic GPT) model, our method consistently improves all six evaluation metrics. 
This result shows that our multi-agent framework brings a significant improvement compared to using GPT directly.
We also compare the performance of GPT-3.5 and GPT-4 for the standard prompting method in clinical trial outcome prediction. As observed in Table \ref{table:result}, GPT-4 demonstrates a superior performance across most metrics compared to GPT-3.5, suggesting that newer versions of large language models may offer incremental improvements in predicting clinical trial outcomes using the standard prompting method.

At the same time, we also observed that another critical metric, PR-AUC,
while showing significant improvement with \mname~compared to GPT-Direct method, 
still lags behind traditional machine learning models. 
This is particularly important in the healthcare domain, where PR-AUC may hold greater significance. 
It is noteworthy that in clinical applications, all machine learning models, including HAtten, can be integrated as external tools within the \mname~framework. In this study, we did not include HAtten as an external tool to focus on comparing the results and highlighting the potential of \mname.

\subsection{Case Study} 

We analyze a realistic case study in Table~\ref{tab:casestudy} (NCTID: NCT00311402).
The trial focused on evaluating the treatment effect of the Aggrenox capsule on cerebrovascular accidents. Aggrenox is a popular drug and contains a combination of aspirin and dipyridamole. First, the user describes the problem using natural language: ``I have designed a clinical trial and hope you can help me predict whether this trial can pass'' and attach the drug name, disease name, and inclusion/exclusion criteria as features. Then, the Planning Agent decomposes the whole problem into three subproblems based on clinical knowledge. It was enhanced by few-shot learning, which gives some representative examples to the GPT as the prompt. 
These three subproblems are clinical trial enrollment, Aggrenox's safety to human bodies, and Aggrenox's efficacy in treating cerebrovascular accidents, which are handled by enrollment agents, safety agents, and efficacy agents, respectively. 
These agents of subproblems can be solved with the help of external tools. For example, we train an enrollment success prediction model and the estimated enrollment failure probability is 0.359. Enrollment Agent automatically recognizes and call the predictive AI model and insert the results (0.359) into the text. Similarly, the Safety Agent (drug safety) identifies that the historical failure rate of Aggrenox capsules is 100\%, indicating its high risk. 
Combining this information, the Reasoning Agent will make the final decision that the trial is highly likely to fail, which is correctly forecasted.

\subsection{Ablation Study}

\subsubsection{Different Versions of GPT}

In this ablation study, we assess the effectiveness of two iterations of the Generative Pre-trained Transformer: GPT-3.5 and GPT-4. Our objective is to explore their performance in the context of predicting outcomes in clinical trials using a standard prompting approach. The comparative analysis focuses on key performance metrics, as detailed in the subsequent table.

The results, as summarized in Table \ref{table:result}, distinctly illustrate that GPT-4 outperforms GPT-3.5 across the majority of evaluated metrics. This enhancement in performance with GPT-4 underscores the potential benefits of integrating more advanced versions of large language models in the domain of clinical trial outcome prediction.

\subsubsection{Impact of Few-Shot Learning}

This segment of the ablation study investigates the influence of few-shot learning techniques when integrated into \mname, a multi-agent framework employing large language models (LLMs). Our comparative analysis pits the version of the model that incorporates few-shot learning against a baseline version that does not utilize these adaptations. The comparative results are encapsulated in the table below.

Table \ref{table:few_shot_ablation} reveals that while the accuracy and F1 score marginally favor the model without few-shot learning, the few-shot adapted model (\mname) exhibits superior performance in terms of ROC-AUC and PR-AUC. This indicates that the incorporation of few-shot learning significantly bolsters the model's proficiency in accurately classifying positive instances, despite a minor trade-off in overall accuracy and precision-recall balance.

\section{Discussion}
This study introduces the groundbreaking Multi-Agent Clinical Trial Helper (\mname), a multi-agent framework that synergizes the advanced capabilities of GPT-4 with sophisticated agent architectures and cutting-edge reasoning technologies like LEAST-TO-MOST and ReAct. Our system significantly enhances the performance of large language models (LLMs) in clinical settings, managing complex trial processes and introducing novel functionalities such as predictive analytics, comprehensive failure analysis, and precise trial duration estimations.

Our evaluations, which include computational benchmarks and expert feedback, underscore the efficiency and effectiveness of \mname~in improving clinical trial outcomes. This integration of LLMs with multi-agent systems not only manages the complexities inherent in clinical trials but also bridges the gap between conversational AI and actionable intelligence in healthcare. \mname~establishes a new benchmark in the application of LLMs to clinical trials, promising a future where advanced AI tools play a crucial role in advancing medical research and patient care.

\paragraph{Limitations}
While the \mname~demonstrates the capability to automatically recognize and decompose user issues, directing them to specialized agents for resolution, it still relies significantly on human intervention for its design and configuration. This dependency on manual input for agent creation limits the system's scalability and adaptability, particularly in dynamic environments where user requirements and contexts evolve rapidly. Further development could focus on integrating machine learning techniques to enable the \mname~to learn from interactions and autonomously update its problem-solving strategies, thereby reducing the need for frequent human oversight and redesign.

These points underscore the need for continuous research and development to fully realize the potential of AI-driven clinical trials, addressing both technical and clinical implications.

\bibliographystyle{ACM-Reference-Format}
\bibliography{custom}

\appendix

\section{Supplementary Material}
\subsection{Enrollment Model}
\label{app:enroll}
\paragraph{Model Architecture}
The architecture of the model consists of the following components 
\begin{itemize}
    \item A transformer encoder layer that processes embeddings of inclusion and exclusion criteria, drugs, and diseases. This layer is designed to capture interactions across these different types of information, crucial for understanding the complexity of trial eligibility.
    \item A fully connected layer that maps the high-dimensional features from the transformer encoder to a single output, indicating the probability of successful enrollment.
    \item A sigmoid activation function applied to the output of the fully connected layer, converting it into a probability measure.
\end{itemize}

\paragraph{Data Processing and Feature Extraction}
For each trial, the Enrollment Agent performs the following steps:
\begin{itemize}
    \item \textbf{Criteria Segmentation:} The agent first segments the eligibility criteria into inclusion and exclusion categories. This segmentation allows for targeted analysis of factors that either qualify or disqualify potential participants.
    \item \textbf{Embedding Generation:} Using the pre-trained BioBERT model, the agent converts text data from the criteria, drugs, and diseases into dense vector representations. These embeddings capture deep semantic features that are essential for accurate model predictions.
    \item \textbf{Feature Aggregation:} The embeddings are then aggregated and fed into the transformer encoder, which processes them to capture the complex dependencies among the criteria, drugs, and diseases.
\end{itemize}

\paragraph{Model Training and Evaluation}
The model is trained on a dataset comprising historical trial data, where each record includes the eligibility criteria, associated drugs and diseases, and the trial outcome regarding enrollment success. The training process involves:
\begin{itemize}
    \item Balancing the dataset to handle disparities in the number of successful and unsuccessful enrollments.
    \item Employing a cross-entropy loss function adjusted for class imbalance, ensuring that the model accurately learns from both positive (successful enrollment) and negative (unsuccessful enrollment) examples.
    \item Evaluation on the test dataset yielded an ROC-AUC score of 0.7037 and an accuracy of 0.7689, indicating effective class differentiation and prediction accuracy. This model exemplifies the integration of advanced Natural Language Processing (NLP) and neural architectures to improve clinical trial design and efficiency.
\end{itemize}

\subsection{Function Calling Definitions}

\label{app:func}
The function for retrieving drug information from the DrugBank database is defined as follows:
\begin{verbatim}
{
  "type": "function",
  "function": {
    "name": "retrieval_drugbank",
    "description": "Retrieves information about a drug 
    from the DrugBank database using the drug's 
    name as input.",
    "parameters": {
      "type": "object",
      "properties": {
        "drug_name": {
          "type": "string",
          "description": "The name of the drug."
        }
      },
      "required": ["drug_name"]
    }
  }
}
\end{verbatim}

The function for retrieving the path connecting the drug to the disease from the Hetionet Knowledge Graph is defined as follows:

\begin{verbatim}
{
    "type": "function",
    "function": {
        "name": "retrieval_hetionet",
        "description": "
         Given the names of a drug and a disease, 
           the model retrieves the path connecting the drug 
           to  the disease from the Hetionet Knowledge Graph.
            Hetionet is a comprehensive knowledge graph that 
            integrates diverse biological information by 
            connecting genes, diseases, compounds, 
            and more into an interoperable framework. 
            It structures real-world biomedical data into 
            a network, facilitating advanced analysis and 
            discovery of new insights into disease mechanisms, 
            drug repurposing, and the genetic underpinnings 
            of health and disease.",
        "parameters": {
            "type": "object",
            "properties": {
                "drug_name": {
                    "type": "string",
                    "description": "The drug name",
                },
                "disease_name": {
                    "type": "string",
                    "description": "The disease name",
                }
            },
            "required": ["drug_name", "disease_name"],
        },
    }
}
\end{verbatim}

\end{document}